\documentclass[runningheads]{llncs}
\usepackage[T1]{fontenc}
\usepackage{graphicx}
\usepackage{makecell}

\begin{document}
\title{How Reliable and Stable are Explanations of XAI Methods?}
%
%
\author{José Ribeiro\inst{1,2,3}\orcidID{0000-0002-8836-4188}\and
Lucas Cardoso\inst{2,3}\orcidID{0000-0003-3838-3214} \and
Vitor Santos\inst{3}\orcidID{0000-0002-7960-3079} \and
Eduardo Carvalho\inst{3}\orcidID{0000-0001-9999-4313} \and
Níkolas Carneiro\inst{3}\orcidID{0000-0002-5097-0772} \and
Ronnie Alves\inst{2,3}\orcidID{0000-0003-4139-0562}
}

\authorrunning{J. Ribeiro et al.}
%
\institute{Federal Institute of Pará - IFPA. Ananindeua, 67125-000, Brazil \and
Federal University of Pará - UFPA. Belém, 66075-10, Brazil \and
Vale Institute of Technology - ITV, Belém, 66055-090, Brazil
\\E-mail:\email{jose.ribeiro\@ifpa.edu.br,  lucas.cardoso@icen.ufpa.br, vitor.cirilo.santos@itv.org, eduardo.costa.carvalho@itv.org, nikolas.carneiro@itv.org, ronnie.alves@itv.org}
}
\maketitle              
\begin{abstract}
Black box models are increasingly being used in the daily lives of human beings living in society. Along with this increase, there has been the emergence of Explainable Artificial Intelligence (XAI) methods aimed at generating additional explanations regarding how the model makes certain predictions. In this sense, methods such as \textit{Dalex}, \textit{Eli5}, \textit{eXirt}, \textit{Lofo} and \textit{Shap} emerged as different proposals and methodologies for generating explanations of black box models in an agnostic way. Along with the emergence of these methods, questions arise such as ``\textit{How Reliable and Stable are XAI Methods?}''. With the aim of shedding light on this main question, this research creates a pipeline that performs experiments using the diabetes dataset and four different machine learning models (LGBM, MLP, DT and KNN), creating different levels of perturbations of the test data and finally generates explanations from the \textit{eXirt} method regarding the confidence of the models and also feature relevances ranks from all XAI methods mentioned, in order to measure their stability in the face of perturbations. As a result, it was found that \textit{eXirt} was able to identify the most reliable models among all those used. It was also found that current XAI methods are sensitive to perturbations, with the exception of one specific method.

\keywords{Explainable Artificial Intelligence  \and Reliable \and Item Response Theory \and Machine Learning.}

\end{abstract}
\section{Introduction}
Technology is increasingly evolving on different fronts and efforts, today Artificial Intelligence is a reality in everyday life in our society. There are many real-world problems that machine learning algorithms seek to solve, making human life more automated, intelligent and less complicated \cite{ghahramani2015probabilistic}.


Black-box models are driven by algorithms that, despite normally presenting high performance scores when faced with classification or regression problems, are not capable of self-explaining their predictions. Transparent models are algorithms that, despite normally presenting lower performance scores when faced with classification and regression problems, are capable of self-explaining their predictions through visual structures created by themselves \cite{molnar2020interpretable}.

With the growing need for models that present high performance --- which implies low transparency \cite{arrieta_explainable_2019_20} --- even in sensitive contexts, there is an increasing demand for methods or tools that can provide information about local explanations (explanation of feature relevance generated around each data instance) and global explanations (when it is possible to understand the logic of all instances of the generated model globally) as a means of making predictions more easily interpretable and also more reliable by humans \cite{darpa_2019}.

The terminologies ``feature relevance ranking'' and ``feature importance ranking'' are widely used as synonyms in the computing community, but have different definitions herein, as shown in \cite{arrieta_explainable_2019_20}.  Feature rankings are regarded as ordered structures whereby each feature of the dataset used by the model appears in a position indicated by a score. The main difference being that, in relevance ranking, the calculation of the score is based on the model output, whereas to calculate the importance ranking of features, the correct label to be predicted is used \cite{arrieta_explainable_2019_20}.

In this context, methods like \textit{eXirt} \cite{ribeiro_exirt}, \textit{Dalex} \cite{dalex_book}, \textit{Eli5} \cite{eli5_ref}, \textit{Lofo} \cite{lofo_ref} , \textit {Shap} \cite{kernel_shap_ref} and \textit{Skater} \cite{skater_ref} emerged to promote the creation of model-agnostic (method that does not depend on the type of model to be explained) and model-specific explanations (method depend of specific type of model) \cite{khan2022model_specific}.


The \textit{eXirt} method was recently developed and published in \cite{ribeiro_exirt}, is based on Item Response Theory (IRT) and its properties. It generates model explanations through feature relevance ranking and information based on IRT properties, allowing a human individual to gain confidence in the model.

Defending the idea of an XAI method capable of explaining machine learning models, while also generating evidence regarding the reliability of this model, such as \textit{eXirt}, is not an easy task. Thus, this article sheds light on how the IRT used by \textit{eXirt} is capable of indicating whether specific models are reliable or not, also showing how stable current XAI methods are. Seeking to answer two main hypotheses: ``\textit{Are the most reliable models, according to the eXirt method, those being less affected by data perturbations?}'' and ``\textit{Do existing XAI methods generate stable explanations even after data perturbations?}''.

Aiming to answer the questions above, this research uses in its analyzes the dataset \textit{diabetes} \cite{diabetes} to build 4 different types of machine learning models, making predictions in the models using test data with perturbations and without perturbation, it creates explanations of these models and finally analyzes the features relevance ranks of the models and the parameters that indicate which models are more reliable, this last analysis generated exclusively by \textit{eXirt}.

The main contributions of this article to the computing community focused on machine learning are: Evaluating IRT methodology as a reliable strategy for XAI methods; Benchmark comparison of state-of-the-art XAI methods exploring stability of models explainabilities; Providing a visual tool to explore reliable of XAI methods through the utilization of IRT's Item Characteristic Curves (ICC).

\section{Related works}

This study conducted an literature review, aiming to identify research that proposes existing XAI methods. This allowed for the identification of the main XAI techniques specifically designed to generate global feature relevance rankings, both in a model-agnostic and model-specific manner, applicable to tabular data.

As a result, a total of six XAI methods were found to be properly validated and compatible with one another (at library and code execution dependencies  level). These methods include: \textit{eXirt} \cite{ribeiro_exirt}, \textit{Dalex} \cite{dalex_python_ref}, \textit{Eli5} \cite{eli5_ref}, \textit{Lofo} \cite{lofo_ref}, \textit{SHAP} \cite{kernel_shap_ref} and \textit{Skater} \cite{skater_ref}.

This survey found other tools aimed at model explanation, including: \textit{Alibi} \cite{alibi_ale_ref}, \textit{CIU} \cite{ciu_ref}, \textit{Lime} \cite{lime_ref}, \textit{IBM Explainable AI 360} \cite{ibm_xai360}, \textit{Anchor} \cite{anchors_aaai18}, \textit{Attention} \cite{lin2017structured_attention} e \textit{Interpreter ML} \cite{interpretML_arxiv}. However, due to incompatibilities and technical problems, they ended up not being used in this research.

The primary issues and incompatibilities identified were: absence of global rank generation; rank generation dependent on another existing XAI method within the pipeline; incompatibility with pipeline dependencies at the library version level; and outdated method libraries.

Note that the six methods presented herein generate relevance rankings based on the same previously trained machine learning models (with the same training and testing split), manipulate their inputs and/or produce new intermediate models copies. Therefore, they are required to be compatible with each other so that a fair comparison of their final rankings of explanations can be made. Table \ref{tab_resumo_xai} shows a general comparison between the techniques found during bibliographic research.


\begin{table*}[!h]
\vspace{-5mm}
\centering
\caption{Overall view of XAI methods}
\resizebox{.95\textwidth}{!}{%
\begin{tabular}{c|c|c|c|c|c|c}
\hline
\textbf{\begin{tabular}[c]{@{}c@{}}Name\end{tabular}} & 
\textbf{Base algorithm} &
\textbf{\makecell{Explanation\\ technique}} & 
\textbf{\makecell{Global \\ explanation\\ (by rank)}} &
\textbf{\makecell{Local \\ explanation}} &
\textbf{\makecell{Model\\ Specific or \\ Agnostic?}} &
\textbf{\makecell{Compatible?}} \\ \hline
\textit{Alibi} &  \makecell{Out-of-bag\\error} & \makecell{Feature\\Permutation\\and accuracy and f1} & Yes & Yes & Agnostic & No \\ \hline
\textit{Anchor} &  \makecell{if-Then\\Rules} & \makecell{Rules} & No & Yes & Agnostic & No \\ \hline
\textit{Attention} &  \makecell{Structured\\Self-attentive\\embedding} & \makecell{Multiple\\Vector\\Representations} & No & Yes & Specific & No \\ \hline
\textit{CIU} &  \makecell{Decision\\Theory} & \makecell{Feature Permutation \\and Multiple Criteria\\Decision Making} & \makecell{Yes\\(deprecated)} & No & Agnostic & No \\ \hline
\textit{Dalex} & \makecell{Leave-one\\ covariate out} & Feature Permutation & Yes & Yes & Agnostic & Yes \\ \hline
\textit{Eli5} & \makecell{Assigning \\weights\\ to decisions} & \makecell{Feature Permutation\\and Mean Decrease\\ Accuracy} & Yes & Yes & Specific & Yes \\ \hline
\textit{eXirt} & \makecell{Item\\ Response\\ Theory} & \makecell{Feature Permutation\\and Model Ability} & Yes & Yes & Specific & Yes \\ \hline
\textit{\makecell{IBM\\Explainable\\AI 360}} &  \makecell{Same of \textit{Shap}} & \makecell{Same of \textit{Shap}} & Yes & Yes & Specific & No \\ \hline
\textit{\makecell{Interpreter\\ML}} &  \makecell{Same of\\\textit{Lime} and \textit{Shap}} & \makecell{Same of\\\textit{Lime} and \textit{Shap}} & Yes & Yes & \makecell{Specific and\\ Agnostic} & No \\ \hline
\textit{Lime} &  \makecell{local linear\\approximation} & \makecell{Perturbation of\\the Instance} & No & Yes & Agnostic & No \\ \hline
\textit{Lofo} & \makecell{Leave One\\ Feature Out} & Feature Permutation & Yes & No & Specifc & Yes \\ \hline
\makecell{\textit{Shap}\\(Kernel)} & \makecell{Game\\Theory} & Feature Permutation & Yes & Yes & Agnostic & Yes \\ \hline
\textit{Skater} & \makecell{Information\\ Theory} & Feature Relevance & Yes & Yes & Agnostic & Yes \\ \hline
\end{tabular}
}
\vspace{-5mm}
\label{tab_resumo_xai}
\end{table*}

In previous studies, this research used the CIU in its tests, however when carrying out this study it was found that its creators updated their libraries and apparently this method no longer generates feature relevance rankings as previously, which is why it appears as incompatible. Still in table \ref{tab_resumo_xai}, it can be seen that most existing XAI methods use the ``Feature Permutation'' technique to perform the model explanation process. However, it should be emphasized at this moment, that the \textit{eXirt} differs from other methods by having base in \textit{IRT}.

Note, although \textit{eXirt} is understood as a model-specific method, table \ref{tab_resumo_xai}. In the research described here, it will be tested with models that go beyond the tree-ensemble, given that its architecture is generalist, as are the two other model-agnostic methods, as mentioned in \cite{ribeiro_exirt}.

\section{Background}

\subsection{Explainable Artificial Intelligence}


Tools like \textit{Dalex}, \textit{Eli5}, \textit{eXirt}, \textit{Lofo}, \textit{SHAP}, and \textit{Skater} can generate various types of explanations. However, for quantitative comparisons, only their ranking generation process are described. To clarify how each method generates feature relevance explanations, their basic operations are detailed below.

The \textit{eXit} is the newest XAI method aimed at creating model explanations from feature relevance ranks based on Item Response Theory. This theory is the same one used in the evaluation of candidates taking tests such as the National Secondary Education Examination from Brazil (in portuguese is ENEM). Making a quick analogy, in the \textit{eXirt} method, the dataset is considered a proof, the model is considered a candidate to answer the proof, the instances of the dataset are the questions, the value of the features of the instances are considered the commands of the questions and the objective value is considered the answer to each question. The \textit{eXirt} based on IRT can evaluate candidates or models through 3 different perspectives: difficulty, discrimination and guessing, generating feature relevance rankings \cite{ribeiro_exirt}.

\textit{Dalex} is a set of XAI tools based on the \textit{LOCO} (Leave One Covariate Out) approach. It receives the model and data, calculates model performance, performs new training with modified datasets, iteratively inverts each feature, and evaluates model performance based on these inversions to identify important features \cite{dalex_book}.

\textit{Leave One Feature Out (Lofo)} is similar to \textit{Dalex} but removes features iteratively instead of inverting them. It evaluates model performance with all features, removes one feature at a time, retrains the model, and assesses performance on a validation dataset, reporting the mean and standard deviation of each feature's relevance \cite{lofo_ref}.

\textit{Eli5} uses the \textit{Mean Decrease Accuracy} algorithm to rank feature relevance by measuring performance decline when a feature is removed from the test dataset \cite{eli5_ref}.

\textit{SHapley Additive exPlanations (SHAP)} explains a prediction by calculating the contribution of each feature, based on game theory’s \textit{Shapley Value}. Features are iteratively included and excluded from models to compute their \textit{Shapley Values}, generating a relevance ranking \cite{kernel_shap_ref}.

\textit{Skater} calculates feature relevance based on Information Theory, measuring entropy changes in predictions when a feature is perturbed. Although now closed source, it remains popular in the XAI community \cite{skater_ref}.

\subsection{Item Response Theory}

Item Response Theory (IRT), part of Psychometrics, provides mathematical models to estimate latent traits, relating the probability of a specific response to the characteristics of the items evaluated. Traditional assessment methods measure performance by the total number of correct answers, but have limitations, such as dealing with random answers and evaluating the difficulty of each test question. \cite{de2000teoria_irt_ref}.

Unlike traditional assessments, IRT focuses on test items, evaluating performance based on the ability to get specific items correct, not just the total count of correct answers \cite{hambleton1991fundamentals_irt_2}. IRT seeks to evaluate unobservable latent characteristics of an individual, relating the probability of a correct answer to their latent traits, that is, to the individual's ability in the area of knowledge evaluated. \cite{cardoso2020decoding_irt}.

In summary, IRT consists of mathematical models that represent the probability of an individual getting an item correct, considering item parameters and the respondent's ability. Different implementations of IRT exist in the literature, such as the ``\textit{Rasch Dichotomous Model}'' \cite{kreiner2012rasch_params_item} and the ``\textit{Birnbaum Three-Dimensional Model}'' \cite{birnbaum1968some_parameters_irt} (the last one used here). In the two topics below, it will be described how the main processes of this theory.

\subsubsection{Estimation of Item Parameters.}

This process involves the estimation of discrimination, difficulty, and guessing based on the 3PL model, using techniques such  \textit{Maximum Likelihood Estimation - MLE}. The objective is find the parameter values that maximize the probability of observing individuals' actual responses to the items, through:

\begin{itemize}
    \item \textit{Discrimination:} consists in how much a specific item $i$ is able to differentiate between highly and poorly skilled respondents. It is understood that the higher its value, the more discriminative the item is. Ideally, a test should feature a gradual and positive discrimination;

    \item \textit{Difficulty:} represents how much a specific item $i$ is hard to be responded correctly by respondents. Higher difficulty values represent more difficult items to answer;
  
    \item \textit{Guessing:} representing the probability that a respondent gets a specific item $i$ right randomly. It can also be understood as the probability that a respondent with low ability will get the item right. It is also the smallest possible chance that an item will be correct regardless of the estimated ability of the respondent.
    
\end{itemize}

\subsubsection{Estimation of ability.}

This process is represented of logistic model $3PL$, presented in the equation \ref{eq:ml3}, consists of a model capable of evaluating the respondents of a test from the estimated ability ($\theta_{j}$), together with the correct answer probability $P(U_{ij} = 1\mid\theta_{j})$ calculated as a function of the individual skill $j$ and the parameters of the item $i$.

\begin{equation}
    \label{eq:ml3}
    P(U_{ij} = 1\mid\theta_{j}) = c_{i} + (1 - c_{i})\frac{1}{1+ e^{-a_{i}(\theta_{j}-b_{i})}}
\end{equation}

The \textit{3PL} is used to model the relationship between individuals' ability and the likelihood of correctly answering an item on a test. It assumes that the probability of a correct answer depends on three item parameters: item discrimination, item difficulty, and item guessing.

In the equation \ref{eq:ml3} the properties discrimination, difficulty and guessing of the items $i$, are represented respectively by the letters $a_i$,$b_i$, and $c_i$. The $\theta_j$ is the ability of the individual $j$, which is a continuous parameter representing the latent trait being measured. $P(U_{ij} = 1\mid\theta_{j})$ represents the probability of correctly answering item i for an individual $j$ with ability $\theta$.

\begin{figure*}[!h]
\begin{center}
\vspace{-3mm}
\includegraphics[scale=0.22]{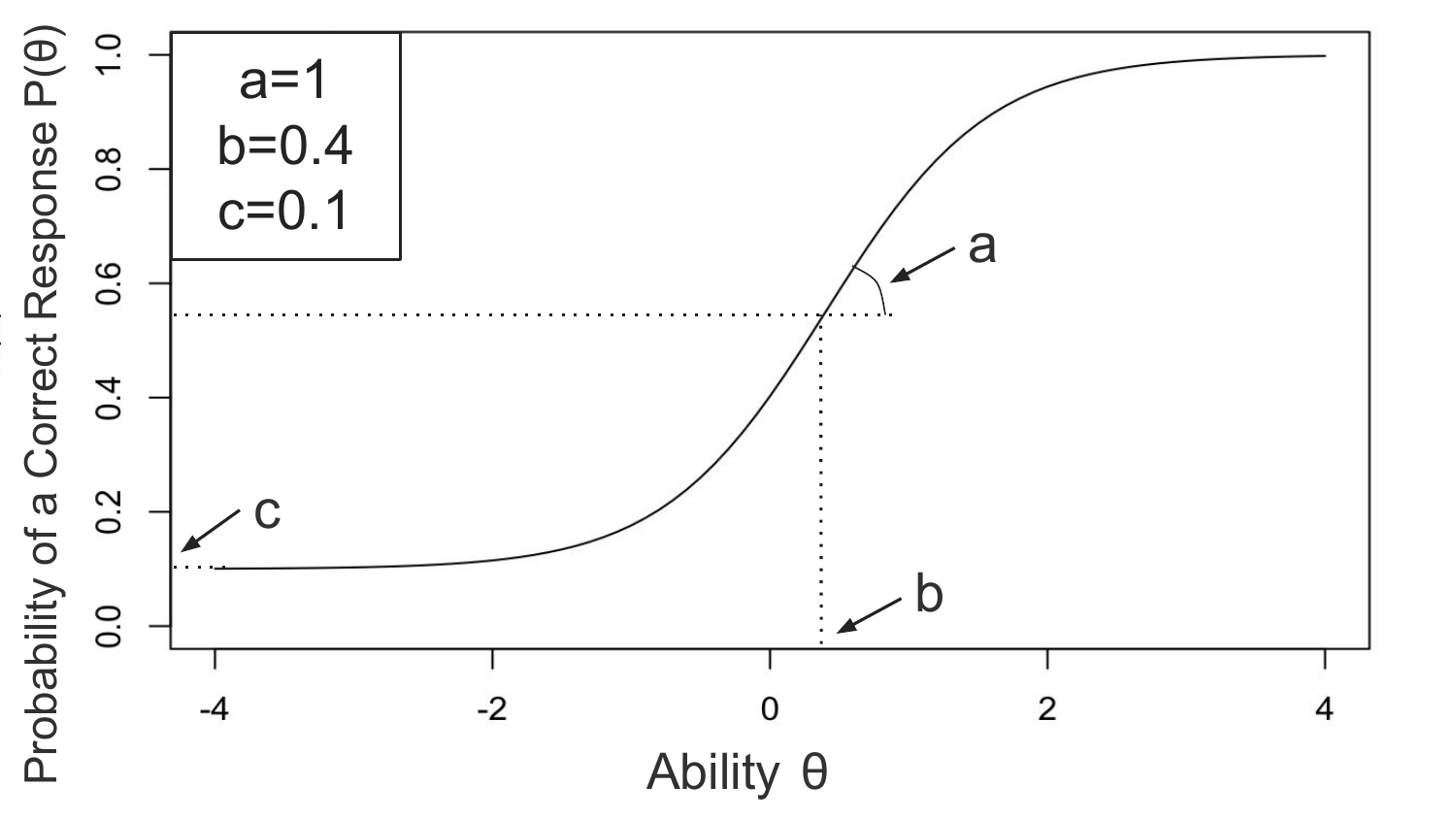}
\caption{The Item Characteristic Curve - ICC, the letters $a$, $b$ and $c$ represent the discrimination, difficulty and guessing properties, respectively.}
\label{fig_cci}
\end{center}
\vspace{-10mm}
\end{figure*}

Thus, once the item parameters are estimated and the hit probability is calculated using the equation \ref{eq:ml3}, the \textit{Item Characteristic Curve (ICC)} can be obtained. The \textit{ICC} defines the behavior of an item's hit probability curve according to the parameters describing the item ($a_i$, $b_i$ e $c_i$) and the respondents' skill variance, figure \ref{fig_cci}.

As can be seen in figure \ref{fig_cci}, the hit probability on axis $y$ is calculated by adding the values of the properties $a_i$, $b_i$ e $c_i$ found in an item and the variation of the skill $\theta$. Thus, the property $a_i$ (discrimination) is responsible for the slope of the logistic curve; the property $b_i$ (difficulty) plots the curve as a function of skill in the logistic function; and the property $c_i$ (guessing) places the basis of the logistic function relative to the axis $y$.

These are basic foundations of IRT that will allow a better interpretation of the explanations generated by the XAI method \textit{eXirt}, more information about the theory in question can be accessed from the study \cite{ribeiro_exirt}.

\subsubsection{Model confidence from \textit{IRT}.}


The conceptual bases of IRT \cite{cardoso2020decoding_irt,cardoso2022explanationbyexample_irt,de2000teoria_irt_ref}, a model is understood as reliable if it presents: Discrimination with higher values (or perfect discrimination), since high discrimination means that the model has greater probabilities of success even using little skill (Note, a negative discrimination means a problem in the analyzed instance, indicating low confidence in the prediction); Difficulty with lower values, since the lower the difficulty, the better and more reliable this model is; Guessing with lower values, since if a model gets few random predictions right, it means it is more reliable; Skill with higher values, the higher the model's skill, the more reliable it is.
    
\section{Methodology}

\begin{figure}[h]
\vspace{-4mm}
\includegraphics[width=\textwidth]{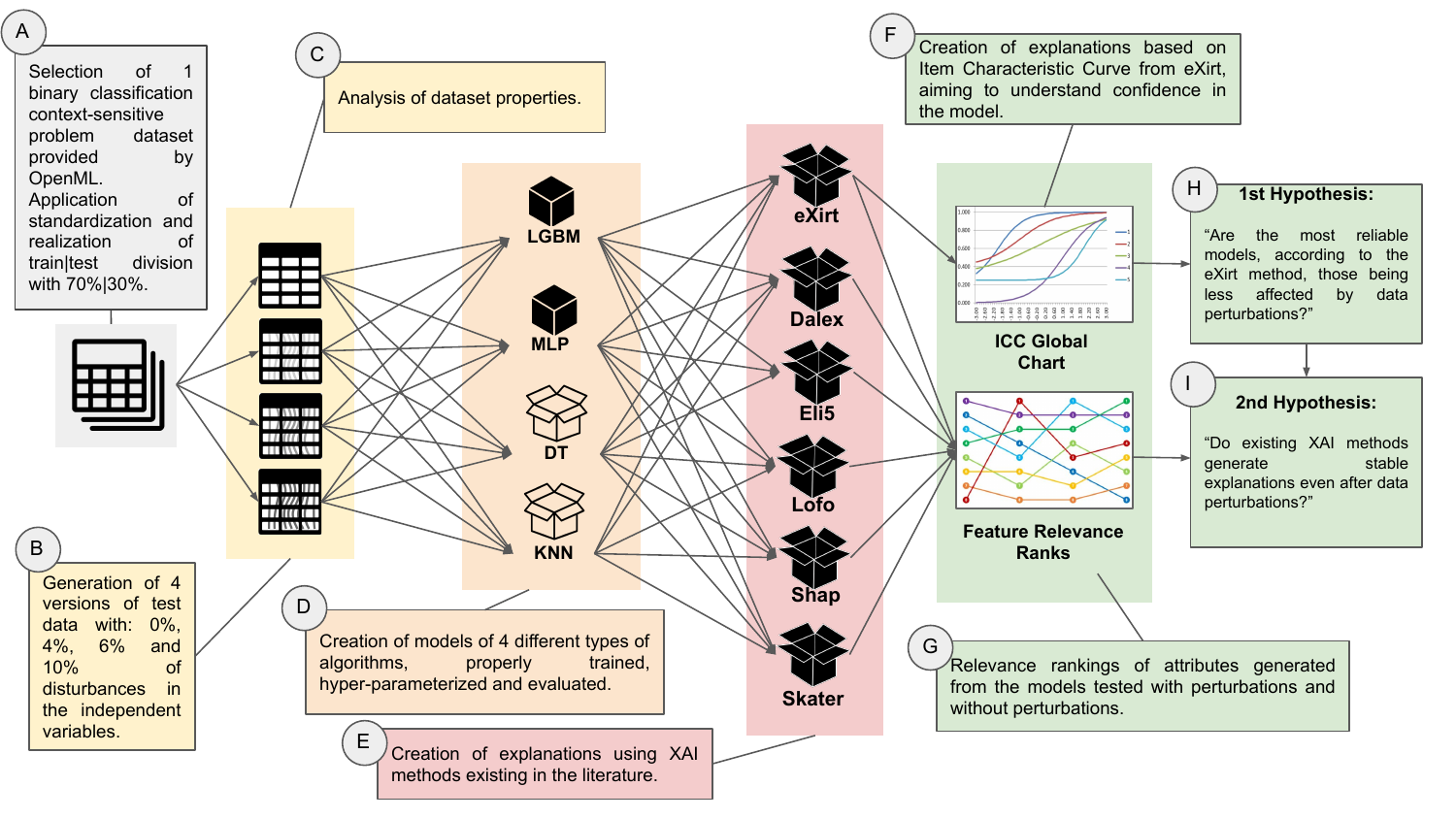}
\caption{Representation of the pipeline of the experiments carried out.}
\label{fig_metodologia}
\vspace{-7mm}
\end{figure}

Aiming to respond to the hypotheses initially launched, a pipeline was developed containing analyzes that can be viewed in the diagram in figure \ref{fig_metodologia}.

The pipeline starts in figure \ref{fig_metodologia} (A), where a dataset was selected relating to a real-world problem (diabetes disease \cite{diabetes}) with a sensitive context (health) defined as binary classification ( aiming to simplify analyses). This database was standardized (using \textit{z score}) and divided between training and testing in the proportion 70\%|30\%.

In figure \ref{fig_metodologia} (C), analyzes of the main properties of the dataset were carried out, as was done in \cite{ribeiro_complexity_et_al_2021}. The diabetes dataset \cite{diabetes} (Pima Indian diabetes dataset) has 9 numeric features entitled: \textit{``Number of times pregnant''}, \textit{``Plasma glucose concentration at 2 hours in an oral glucose tolerance test''}, \textit{``Diastolic blood pressure (mm Hg)''}, \textit{``Triceps skin fold thickness (mm)''}, \textit{``2-Hour serum insulin (mu U/ml)''}, \textit{``Body mass index (weight in kg/(height in m)$^2$)''}, \textit{``Diabetes pedigree function''}, \textit{``Age (years)''}, \textit{``Class variable (0 or 1)''}. It has no missing data and has a 768 instances (500 referring to 0 and 268 referring to 1, classes).

Four different types of models were created in figure \ref{fig_metodologia} (D), which are: Multilayer Perceptron (MLP), Light Gradient Boosting Machine (LGBM), K Nearest Neighbor (KNN) and Decision Tree (DT), all hyper parameterized (with crossvalidation with folds=4, evaluated by Area Under the Curve (AUC)) aiming to improve the adaptability and performance of the models in the process of generalizing the problem represented by the diabetes dataset.

The choice of the types of models was guided by the diversification of tests, including black box models and transparent models, thus LGBM was chosen as a representative of ensemble-based models (black box), the MLP algorithm with a representative of neural networks (black box), the DT algorithm as a representative of tree-based algorithms (transparent) and the KNN algorithm as a representative of algorithms based on instance distance (transparent). Evidently, many other types of models could be used to achieve a greater scope of results, but due to issues related to computational cost, it was decided to use 4 types.

Taking as a basis the test set of the dataset, figure \ref{fig_metodologia} (B), 4 different versions of it were created, containing different percentage levels of instance perturbations: 0\% (original test set), 4\%, 6\%, and 10\%. These percentages were selected from tests with perturbations that varied between 0\% to 40\%, as the focus is to select different test sets that, through perturbations, would succinctly and gradually harm the performance of most models.

Tests were carried out with two types of perturbations, one that applied random noise controlled by percentage in all instances of the data passed to it, as well as in \cite{molnar2020interpretable,robnik2018perturbation_1}. And another form of perturbation that exchanges a specific percentage of the data instances, as done in \cite{chang2018explaining_perturbation_2}. After verifying the drop in performance of the perturbed models, it was decided to use only the permutation, as it proved to be better at perturbing the model aiming at its continuous drop in performance.

Then, XAI techniques were applied (\textit{Dalex, Eli5, eXirt, Lofo, Shap} and \textit{Skater}), figure \ref{fig_metodologia} (E) aiming to create explanations of the models (tested based on of perturbed and unperturbed data). In the stage of figure \ref{fig_metodologia} (G), all models generated explanations based on feature relevance ranks (compared using Bump Chart and Spearman Correlation \cite{spearman_ref}), but aiming to extract information about the confidence of models also generated explanations based on Item Characteristic Curve (ICC), exclusively from \textit{eXirt}, figure \ref{fig_metodologia} (F).

As above, it is clear that only a single dataset was used in the analyses. This fact does not limit the research results, as it is exactly what is intended as the focus of the analyses, as having a single dataset (clean, properly processed and without perturbations) and this data can be considered the most reliable, especially when compared to alternative versions of this data that were perturbated.

However, it is understood that the results collected through the analyzes of this pipeline are sufficient to answer the hypotheses in figure \ref{fig_metodologia} (H) and (I), and can also be generalized to other machine learning contexts. For reproducibility purposes, follow the link to the official pipeline repository of this article 
\url{https://github.com/josesousaribeiro/XAI-eXirt-vs-Trust}.

\section{Results and discussion}

The first step to understanding the pipeline results is to observe the behavior of the models' performances obtained from tests with perturbation and without perturbation. The table \ref{tab_performance} shows the results of the 4 models, as well as the values obtained from the Accuracy, Precision, Recall, F1 and Roc AUC metrics, based on tests with different percentages of perturbations.

\begin{table}[]
\centering
\vspace{-6mm}
\caption{Summary of the model's performances, according to tests carried out.}
\resizebox{.8\textwidth}{!}{%
\begin{tabular}{|r|cccc|cccc|cccc|cccc|}
\hline
\multicolumn{1}{|c|}{-} & \multicolumn{4}{c|}{LGBM}          & \multicolumn{4}{c|}{MLP}           & \multicolumn{4}{c|}{DT}            & \multicolumn{4}{c|}{KNN}           \\ \hline
Perturbation             & \textbf{0\%}  & 4\%  & 6\%  & 10\% & \textbf{0\%}  & 4\%  & 6\%  & 10\% & \textbf{0\%}  & 4\%  & 6\%  & 10\% & \textbf{0\%}  & 4\%  & 6\%  & 10\% \\
Accuracy                & \textbf{0.76} & 0.75 & 0.75 & 0.74 & \textbf{0.74} & 0.57 & 0.58 & 0.57 & \textbf{0.72} & 0.59 & 0.59 & 0.57 & \textbf{0.71} & 0.71 & 0.71 & 0.69 \\
Precision               & \textbf{0.70} & 0.68 & 0.68 & 0.65 & \textbf{0.66} & 0.37 & 0.38 & 0.35 & \textbf{0.65} & 0.38 & 0.38 & 0.35 & \textbf{0.61} & 0.61 & 0.61 & 0.57 \\
Recall                  & \textbf{0.57} & 0.56 & 0.56 & 0.53 & \textbf{0.53} & 0.29 & 0.31 & 0.28 & \textbf{0.48} & 0.28 & 0.28 & 0.26 & \textbf{0.49} & 0.49 & 0.49 & 0.46 \\
F1                      & \textbf{0.63} & 0.61 & 0.61 & 0.58 & \textbf{0.59} & 0.33 & 0.34 & 0.31 & \textbf{0.55} & 0.33 & 0.32 & 0.3  & \textbf{0.55} & 0.55 & 0.55 & 0.51 \\
Roc AUC                 & \textbf{0.72} & 0.71 & 0.71 & 0.69 & \textbf{0.69} & 0.51 & 0.52 & 0.50 & \textbf{0.67} & 0.52 & 0.52 & 0.5  & \textbf{0.66} & 0.66 & 0.66 & 0.63 \\ \hline
\end{tabular}
\label{tab_performance}
}
\vspace{-8mm}
\end{table}

It can be seen in the table \ref{tab_performance} column with 0\% perturbation, that the model with the best performance was LGBM, followed by MLP, DT and KNN. Such results were already expected, as according to \cite{arrieta_explainable_2019_20}, in general, black box models perform better than transparent models. Based on the perturbations added to the test set, it was possible to obtain gradual and succinct worsening of performance in the tests with the increase in perturbations (4\%, 6\% and 10\%).

\begin{figure}[h]  
\vspace{-5mm}
\centering
\includegraphics[scale=0.4]{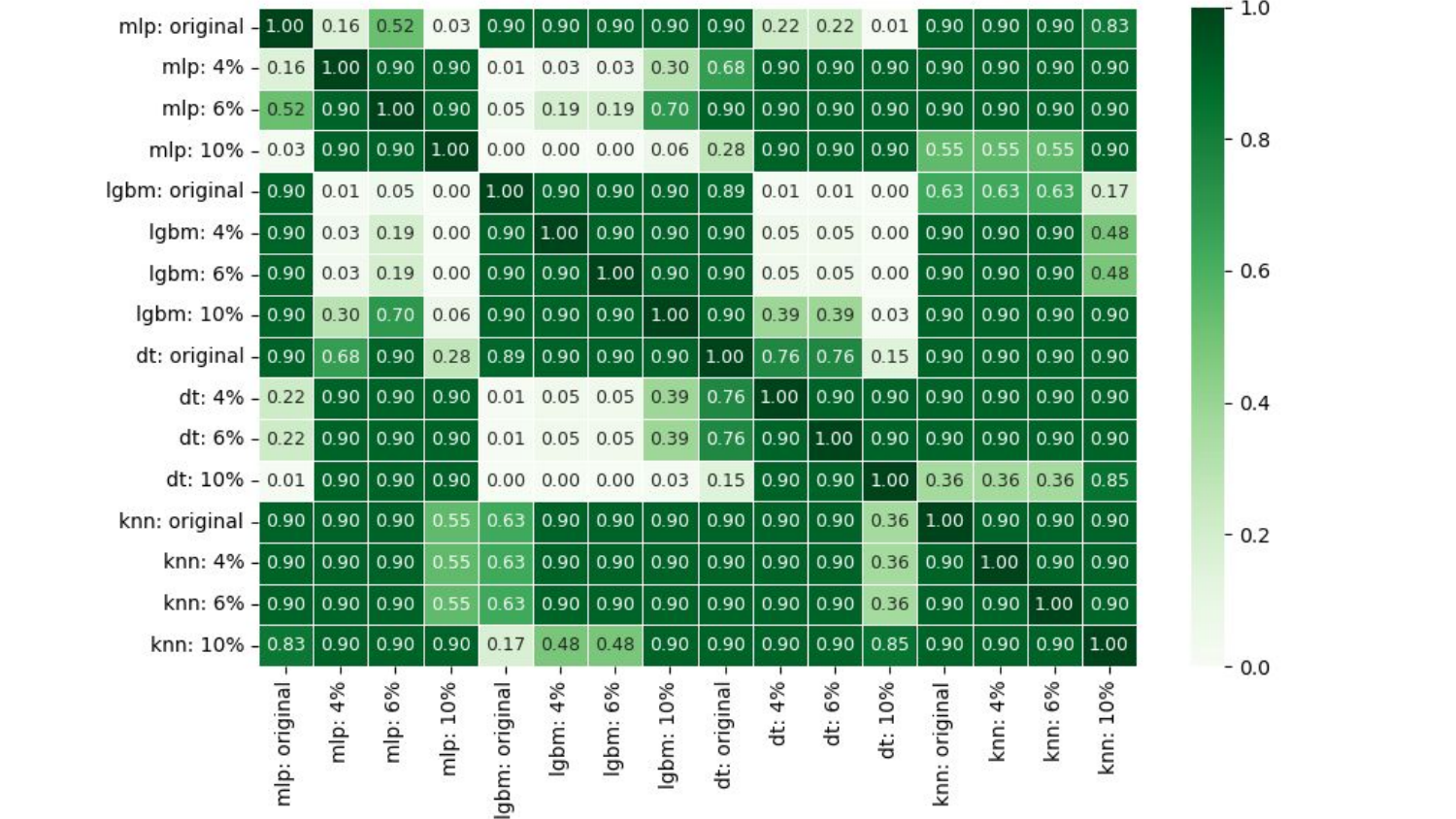}
\label{fig_statistical_test}
\caption{Statistical test summary \textit{Friedman Nemenyi}.}
\vspace{-6mm}
\end{figure}

The percentages of perturbations 4\% and 6\% work as controls in the analyses, as it can be seen in some cases, a larger perturbation makes the model perform minimally better than a smaller perturbation. However, this does not occur with perturbations equal to 10\%. In order to identify how significant the difference in performance of the models is, the statistical test \textit{Friedman Nemenyi} \cite{demvsar2006statistical_compare_mult_classifier_friedman} was carried out, figure \ref{fig_statistical_test}.

In the test presented in figure \ref{fig_statistical_test}, the \textit{p-value} values of \textit{Friedman Nemenyi} are shown relating to the comparisons of the statistical metrics shown in the table \ref{tab_performance}. In each cell of the matrix, the closer the value is to 0 (zero), the greater the statistical confidence that can be obtained. Thus, adopting the value of \textit{p-value}$= 0.05$ as a cutoff, it can be determined with at least 95\% statistical confidence that the models present different performance. Thus, when observing the row ``lgbm: original'' and the columns ``mlp: original'', ``dt: original'' and ``knn: original'', it can be seen that the performances of the models without perturbations do not present a statistically significant difference.

\subsection{Are the most reliable models, according to the \textit{eXirt} method, those being less affected by data perturbations?}

\begin{figure}[h]
\vspace{-5mm}
\includegraphics[width=\textwidth]{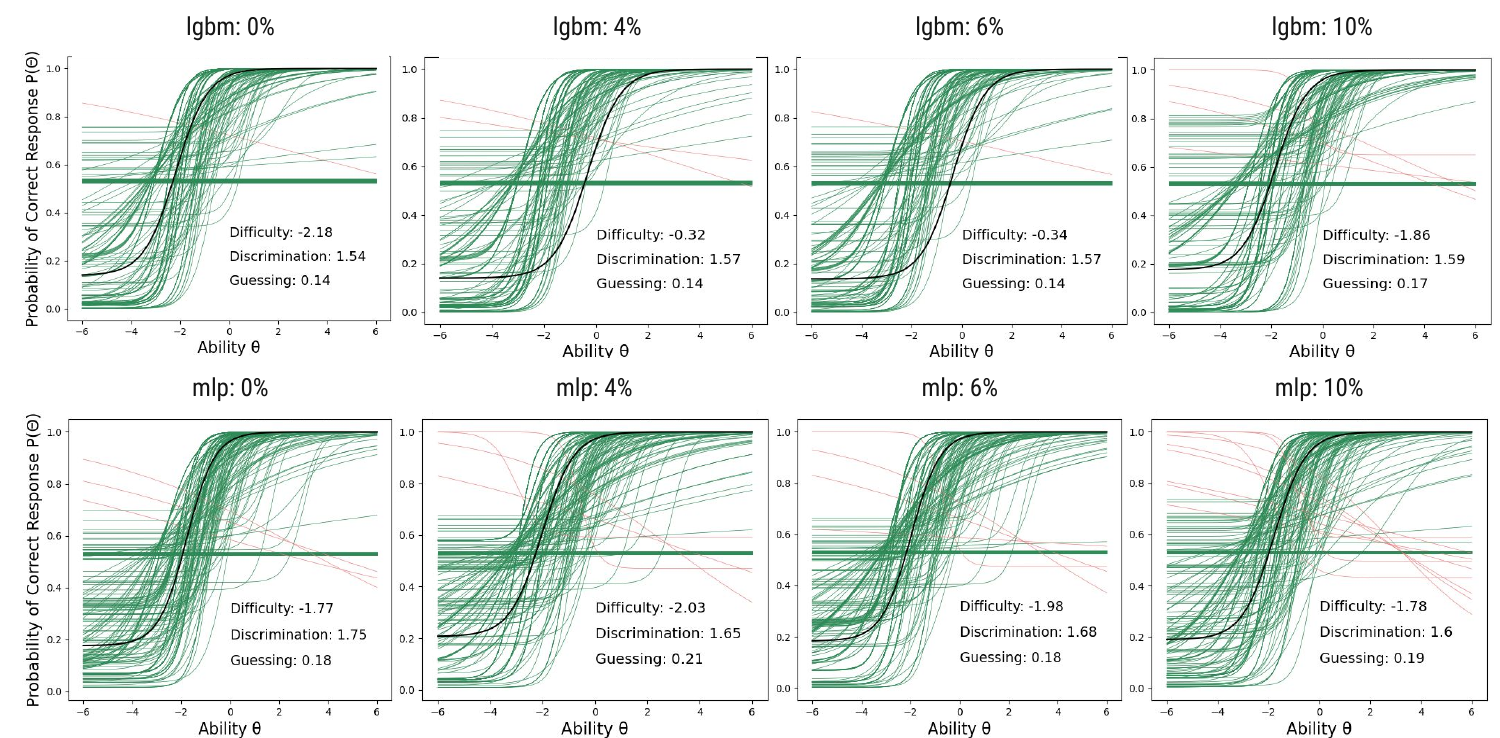}
\label{fig_lgbm_mlp}
\vspace{-5mm}
\caption{Summary of the global ICC of the LGBM and MLP models.}
\vspace{-6mm}
\end{figure}

\begin{figure}[h]
\vspace{-5mm}
\includegraphics[width=\textwidth]{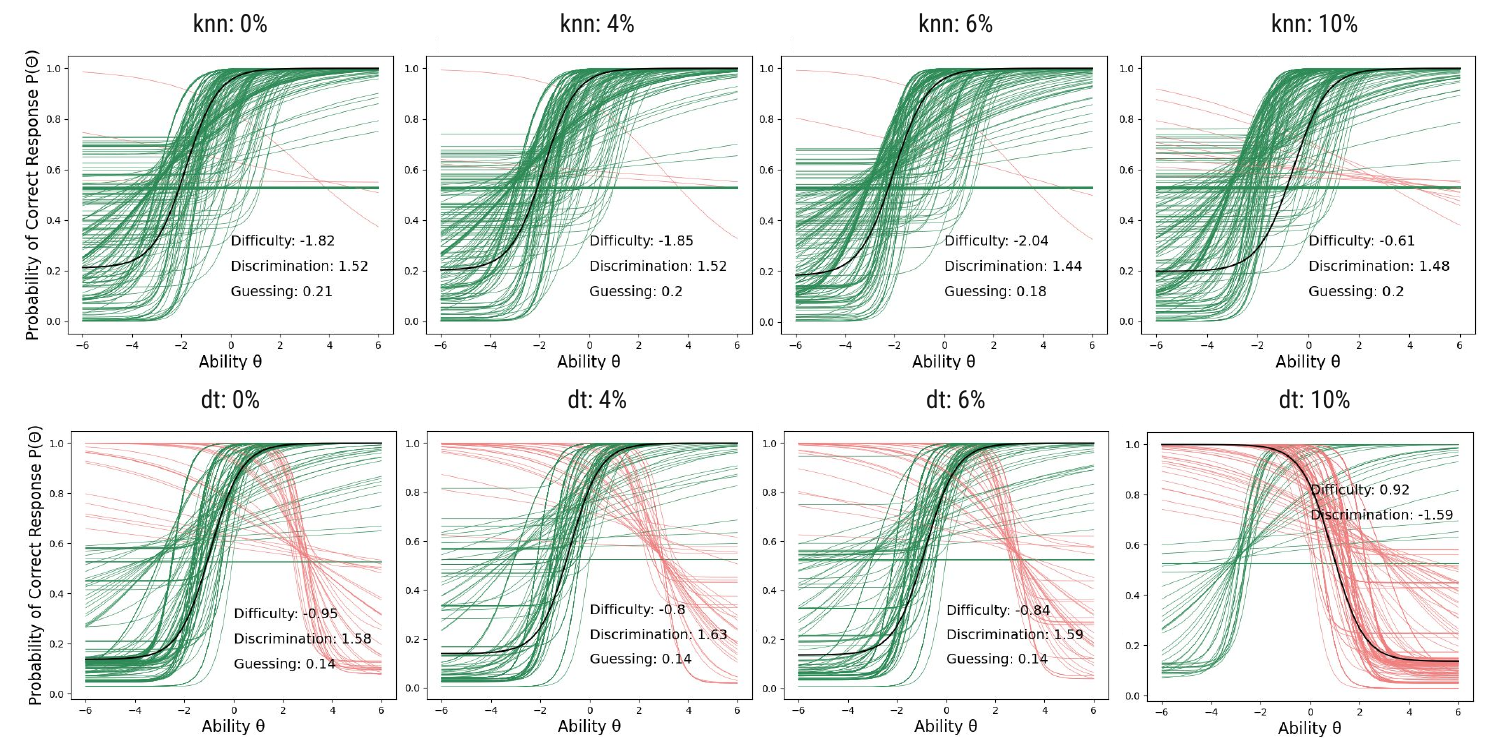}
\label{fig_knn_dt}
\vspace{-5mm}
\caption{Summary of the global ICC of the KNN and DT models.}
\vspace{-6mm}
\end{figure}

The figures \ref{fig_lgbm_mlp} and \ref{fig_knn_dt} were generated exclusively by \textit{eXirt} seeking to answer the first hypothesis released, where, the green and red lines represent the instances of the test dataset that were passed to the model. The black line (thicker) represents the average of the ICCs. The texts with difficulty, discrimination and guessing values present are averages of the curves found.

The experiments that generated the figures \ref{fig_lgbm_mlp} and \ref{fig_knn_dt} follow the line of reasoning: when faced with testing models with instance perturbations and without perturbations, it must be understood that the most reliable models/tests (referring to how can a human trust) are the unperturbed ones. Therefore, it is expected that the results of unperturbed models will be more stable and therefore reliable.

Thus, when observing the figure \ref{fig_lgbm_mlp}, it can be seen that even for models with unperturbed tests (first column on the left of the figure), \textit{eXirt} was able to indicate the most reliable model, in this case the LGBM model, as it presents the least difficulty (-2.18), discrimination (1.54) and lowest guessing value (0.14), compared to the MLP values.

When observing the item characteristic curves generated from the insertion of perturbations, figure \ref{fig_lgbm_mlp}, it can be seen that the difficulty, discrimination and guessing values change succinctly, as in tests with perturbations of 0\% and 10\% of instances, the difficulty goes from -2.18 to -1.86 (LGBM) and from -1.77 to -1.78 (MLP). Discrimination increases from 1.54 to 1.59 (LGBM) and from 1.75 to 1.6 (MLP). Finally the guess goes from 0.14 to 0.17 (LGBM) and 0.18 to 0.19 (MLP). It is noted that the relationships between difficulty and discrimination are inversely proportional, with guessing being proportional to difficulty, not obeying linear proportional rules.

The results in figure \ref{fig_lgbm_mlp} without perturbation show that \textit{eXirt} was able to indicate the most reliable models, as it was able to indicate the best model that obtained the best performance (between MLP and LGBM) and also indicated the unperturbed models/tests as being the most reliable (less difficulty, greater discrimination and less guessing).

Regarding results from the KNN and DT models, figure \ref{fig_knn_dt}, there are some impasses in the process of defining the most reliable model from eXirt, as regarding difficulty there is -1.82 (KNN) and -0.95 (DT), indicating KNN as more reliable. Regarding discrimination, there are 1.52 (KNN) and 1.58 (DT), indicating DT as the most reliable. Finally, the guess value is 0.21 (KNN) and 0.14 (DT), indicating that the model (DT) is the most reliable. In this case, depending on the perspective, one can choose KNN or DT as being the most reliable.

When analyzing the inserted perturbations, figure \ref{fig_knn_dt}, in addition to the gradual worsening of the difficulty, discrimination and guessing values, the numerous appearance of red lines in the ICC graphs stands out. These curves represent negative discrimination, which means that the model identified possible problems in predicting certain records in the dataset, problems related to feature values, that is, as seen in the results of the experiments shown in the figures \ref{fig_lgbm_mlp} and \ref{fig_knn_dt} there are models that are more sensitive and less sensitive to problems such as pertubations in the input data, and \textit{eXirt} showed the DT as the model that suffered most from these perturbations that are related to the nature of the data itself ( since red lines appear in high quantities even when the test has 0\% perturbation).

Note, based on the 12 tests shown in the figures \ref{fig_lgbm_mlp} and \ref{fig_knn_dt}, it can be stated that \textit{eXirt} was able to identify the least reliable models, with higher percentages of perturbations, through the values of the item characteristics. Reinforcing that negative discrimination (red lines in the graphs) were decisive to better characterize the models.

Responding to the hypothesis launched at the beginning of this sub-section, we have the following answer: Yes, since \textit{eXirt} through the properties of difficulty, discrimination and guessing, was able to identify more reliable models (without perturbations) of the less reliable models (with perturbations). It is also able to identify which machine learning models are most reliable, even if these models are of different types and do not present a statistically significant difference in their performance.

\subsection{Do existing XAI methods generate stable explanations even after data perturbations?}

Aiming to answer the hypothesis released, the relevance ranks of features generated from the XAI methods \textit{Dalex}, \textit{Eli5) were selected. }, \textit{eXirt}, \textit{Lofo}, \textit{Shap} and \textit{Skater}, aiming to evaluate their behavior given the need to model explanations with no perturbations and with perturbations. The central idea here is to identify the methods that are most stable to perturbations.

\begin{figure}[!h]
\centering
\includegraphics[scale=0.45]{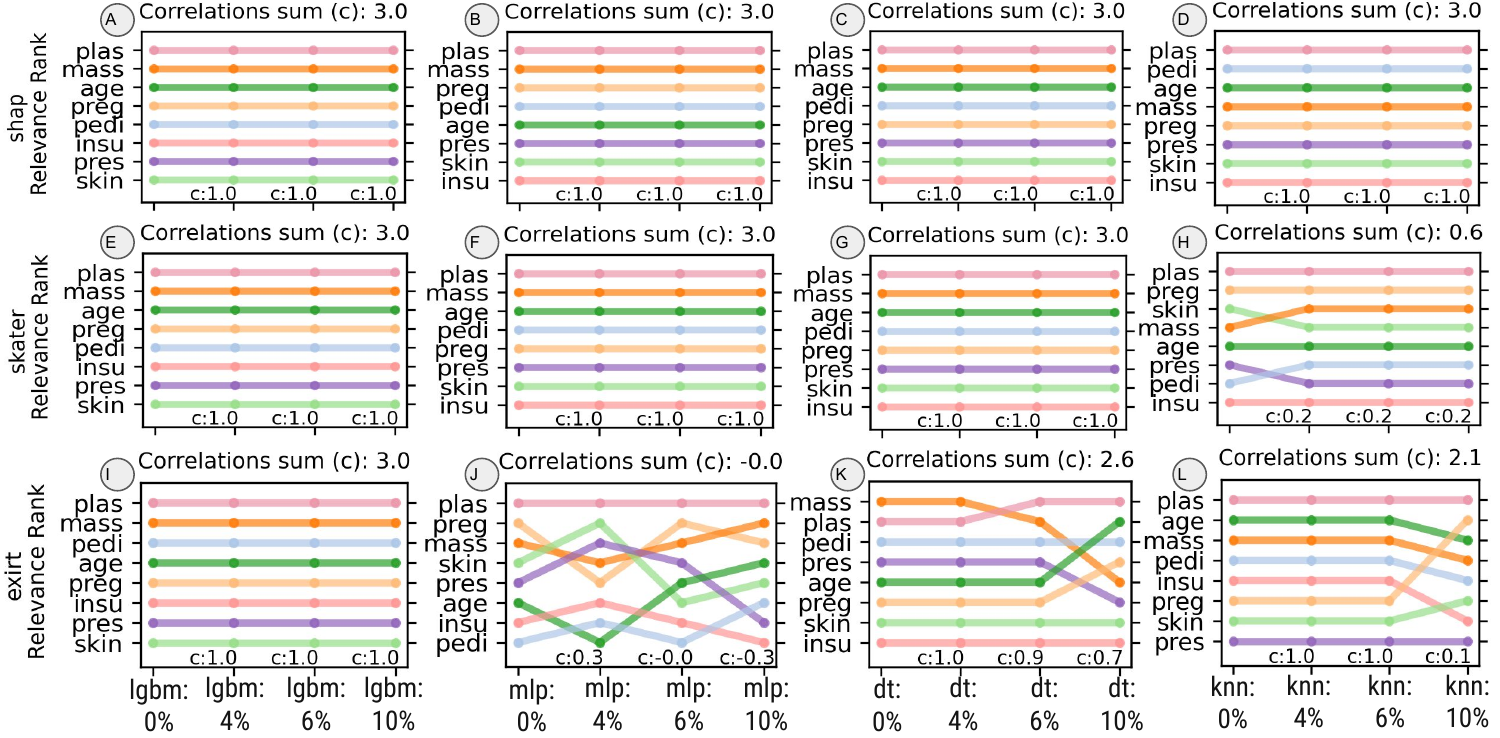}
\label{figs_bumps}
\vspace{-3mm}
\caption{Summary of feature relevance ranks generated in all tests.}
\vspace{-5mm}
\end{figure}

In view of this, we have the figure \ref{figs_bumps} as a general summary that shows all the relevance ranks of features generated from the tests carried out. Each line references an XAI method, ordered from most stable (topmost in the figure) to least stable (bottommost in the figure). Each column references different models (LGBM, MLP, DT and KNN).

Sub-figures referring to the experiments are also shown, where the names of the features present in each rank are indicated on the y axis and the percentages of perturbations are indicated on the x axis: 0\%, 4\%, 6\% and 10\% (from left to right). In these sub-figures, the \textit{Spearman Correlation} values existing between the ranks generated by models with the presence of perturbations in relation to the ranks generated by models without perturbations are also displayed. In the title of these sub-figures the sum of the calculated correlations is presented.


As shown in figure \ref{figs_bumps}, it can be seen that the \textit{shap} method was the most stable XAI method, presenting the same feature relevance rank for the four models, figure \ref{figs_bumps} (A, B, C and D), bearing the maximum value of the sums of correlations (sum = 3) in each model --- with changes being observed in the ranks generated by \textit{shap} only in perturbations above 30\%, experiments carried out separately). Next are the results of the skater method, which presented the same ranks for the LGBM (sum = 3), MLP (sum = 3) and DT (sum = 3) models, generating results with lower correlations for the KNN (sum = 0.57), figure \ref{figs_bumps} (E, F, G and H).

Next comes \textit{eXirt}, which presented equal correlations between the ranks generated from the LGBM model (sum = 3) and high correlations in the explanations of the DT (sum = 2.6) and KNN (2.12) models, figure \ref {figs_bumps} (I,K and L). However, it is worth highlighting the low value of the sum of the correlations found from the MLP model (sum = 0), figure \ref{figs_bumps} (J). Next come the dalex, eli5 and lofo methods, respectively, as methods with lower stabilities in the face of perturbations in the model inputs (full chart: 
\url{https://github.com/josesousaribeiro/XAI-eXirt-vs-Trust/blob/main/output/fig/bump\_ranks.pdf}).

Given these results, one can respond to the hypothesis launched at the beginning of this sub-section, which is: Partially yes, since not all XAI methods tested were capable of generating stable explanations in the face of perturbations in the model inputs. This shows that most XAI methods are currently sensitive to small changes in the way the data is expected in the prediction process, which makes these methods limited in sensitive contexts. Showing evidence that current XAI methods, with the exception of \textit{Shap}, still need to improve their stability in the face of scenarios involving predictions of real data. However, it is noteworthy that \textit{eXirt} differs from other methods because it is capable of generating relevant information about latent characteristics of the model, based on IRT.

\section{Conclusion and Future Works}
This research showed how reliable and stable the explanations of current XAI methods aimed at generating feature relevance ranks are. Highlighting \textit{eXirt} as an XAI method capable of generating extra information regarding how reliable a model is from the IRT perspective (and its properties difficulty, discrimination and guessing). This research also showed that current XAI methods, with the exception of \textit{shap}, are considerably sensitive to changes in model inputs, showing that these methods require greater attention when used in real-world problems.

As future work regarding the research, the following points stand out: Creation of an equation capable of transforming the values of difficulty, discrimination, and guessing properties generated by \textit{eXirt} into a single score that allows for faster interpretation of confidence in the model; Testing \textit{eXirt} with other types of prediction problems such as regression and multiclass classification; Exploring existing XAI methods through new tests and perturbations, aiming to evaluate their behavior in the face of adversity.

\begin{credits}

\subsubsection{\discintname}
The authors declare that they have no conflicting interests with the subjects covered in this research.
\end{credits}
%
%
%
\bibliographystyle{splncs04}
\bibliography{mybib}
\end{document}